\documentclass[conference]{IEEEtran}
\IEEEoverridecommandlockouts
\usepackage{amsmath,amssymb,amsfonts}
\usepackage{textcomp}
\usepackage{xcolor}
\def\BibTeX{{\rm B\kern-.05em{\sc i\kern-.025em b}\kern-.08em
    T\kern-.1667em\lower.7ex\hbox{E}\kern-.125emX}}
\usepackage{graphicx}
\usepackage{hyperref}
\usepackage{xcolor}
\usepackage{booktabs}
\usepackage{amsmath}
\usepackage{graphicx}
\usepackage[caption = false]{subfig}
\usepackage{multirow}
\usepackage[bottom]{footmisc}
\usepackage{footnote}
\makesavenoteenv{tabular}
\makesavenoteenv{table}
\usepackage{cite}
\usepackage{array}
\usepackage[ruled,vlined]{algorithm2e}
\usepackage[most]{tcolorbox}
\usepackage{comment}
\usepackage{soul}
\usepackage{babel,blindtext}
\DeclareMathOperator*{\argmax}{arg\,max}

\usepackage{makecell} 
\usepackage{tabularx}
\usepackage{array}
\newcommand{\PreserveBackslash}[1]{\let\temp=\\#1\let\\=\temp}
\newcolumntype{C}[1]{>{\PreserveBackslash\centering}p{#1}}
\newcolumntype{R}[1]{>{\PreserveBackslash\raggedleft}p{#1}}
\newcolumntype{L}[1]{>{\PreserveBackslash\raggedright}p{#1}}

\begin{document}

\title{FAL-CUR: Fair Active Learning using Uncertainty and Representativeness on Fair Clustering}

\author{\IEEEauthorblockN{Ricky Fajri$^a$\IEEEauthorrefmark{1},
Akrati Saxena$^b$\IEEEauthorrefmark{2}, Yulong Pei$^a$\IEEEauthorrefmark{3} and
Mykola Pechenizkiy$^a$\IEEEauthorrefmark{4}}
\IEEEauthorblockA{$^a$Department of Mathematics and Computer Science \\
Eindhoven University of Technology,
the Netherlands\\
$^b$Leiden Institute of Advanced Computer Science, \\Leiden University, the Netherlands\\
Email: \IEEEauthorrefmark{1}r.m.fajri@tue.nl,
\IEEEauthorrefmark{2}a.saxena@liacs.leidenuniv.nl,
\IEEEauthorrefmark{3}y.pei.1@tue.nl,
\IEEEauthorrefmark{4}mykola.pechenizkiy@tue.nl}}

\maketitle

\begin{abstract}
Active Learning (AL) techniques have proven to be highly effective in reducing data labeling costs across a range of machine learning tasks. Nevertheless, one known challenge of these methods is their potential to introduce unfairness towards sensitive attributes. Although recent approaches have focused on enhancing fairness in AL, they tend to reduce the model's accuracy. To address this issue, we propose a novel strategy, named \textbf{F}air \textbf{A}ctive \textbf{L}earning using fair \textbf{C}lustering, \textbf{U}ncertainty, and \textbf{R}epresentativeness (FAL-CUR), to improve fairness in AL. FAL-CUR tackles the fairness problem in AL by combining fair clustering with an acquisition function that determines which samples to query based on their uncertainty and representativeness scores. We evaluate the performance of FAL-CUR on four real-world datasets, and the results demonstrate that FAL-CUR achieves a 15\% - 20\% improvement in fairness compared to the best state-of-the-art method in terms of equalized odds while maintaining stable accuracy scores. Furthermore, an ablation study highlights the crucial roles of fair clustering in preserving fairness and the acquisition function in stabilizing the accuracy performance.
\end{abstract}


\section{Introduction}
\textcolor{black}{Recently, the world has experienced the advancement of deep neural networks ranging from the natural language processing \cite{Devlin2019BERTPO,GPT-3} to natural image creation \cite{Dall-e}. While showing great results, many deep learning approaches rely on large labeled training data, which is costly and time-consuming to generate. Thus, many works introduced Active Learning (AL) \cite{Settles2009ActiveLL} frameworks to reduce the cost of data annotation. AL methods work by identifying and selecting the most informative or representative samples to be labeled by human experts. These methods have shown great potential to reduce the cost of data annotation, thus increasing the efficiency of the machine learning model. However, preserving fairness in active learning data selection across various population groups remains an unsolved challenge.}

\textcolor{black}{Initially, many works in active learning focused on maximizing accuracy within a given budget. As a result, there is limited research on fair active learning approaches, which seek to reduce disparities in algorithmic performance across underrepresented groups \cite{Anahideh2020FairAL,Sharaf2020PromotingFI}. Furthermore, it has been observed that the performance of active learning methods, often suffers when striving to maintain fairness constraints \cite{Natalia,zhao2019inherent}. The existing fair active learning research, such as FAL \cite{Anahideh2020FairAL,Shen2022MetricFairAL}, shows that when fairness is improved, there is often a noticeable decline in the performance of active learning models. Therefore, there is an open challenge to develop active learning methods that are able to preserve fairness without compromising on accuracy.}


\textcolor{black}{To solve this challenge, we present a novel approach that focuses on fairness in active learning. We present a fair active learning method, called \textbf{F}air \textbf{A}ctive \textbf{L}earning using fair \textbf{C}lustering, \textbf{U}ncertainty and \textbf{R}epresentativeness (FAL-CUR), consisting of two stages. 
In the first stage, we apply a \textit{Fair-Clustering Algorithm} to group uncertain samples while maintaining the fairness constraints. The second stage is the \textit{Selection Algorithm}, which picks out the most informative sample based on fairness ranking measured by the representativeness and uncertainty inside the fair clusters. 
}

The proposed FAL-CUR method solves the problem of selecting data from an unlabeled pool ($\mathcal{U}$) for AL while maintaining fairness constraints, including \textit{Equal Opportunity}, \textit{Equalized Odds}, and \textit{Statistical Parity} metrics. To measure its accuracy and fairness performance, we conducted extensive experiments and in-depth analysis on various real-world datasets. Experimental results show that the FAL-CUR method achieves high fairness performance while maintaining accuracy as compared to baseline active learning models. The ablation analysis shows that all components of the proposed model collectively contribute to achieving high fairness and performance in active learning.

Thus, we summarize the main contributions of our paper as follows.
\begin{itemize}
\item This paper proposes a novel method, FAL-CUR, that is specifically designed to address fairness in active learning procedures.
\item We also present a new active learning acquisition function based on uncertainty and representative score which consistently delivers enhanced performance while improving fairness.
\item Through extensive experiments, the paper shows that the FAL-CUR method yields superior fairness performance compared to existing state-of-the-art fair active learning models on four different datasets.
\end{itemize}

The rest of the paper is organized as follows. Section \ref{sec:related work} introduces the line of works related to this paper. Sections \ref{sec:problem} and \ref{sec:methodology} present the problem definition and the proposed methodology, respectively. Section \ref{sec:expsetup} discusses the experimental setup. In Section \ref{sec:results}, we experimentally investigate the performance of FAL-CUR and elaborate in detail on the performance of FAL-CUR compared to other models. 
Finally, Section \ref{sec:conclusion} concludes the study.

\section{Related Work}\label{sec:related work}
This section summarizes relevant works to our study, including active learning, fair machine learning, and fair active learning. 

\subsection{Active Learning}
The goal of active learning is to reduce the cost of annotation by labeling high-quality samples \cite{Freund2004SelectiveSU, Dasgupta2008HierarchicalSF,Huang2014ActiveLB}. It is very useful in a situation where there is a large amount of training data available, but the cost of annotation is high \cite{Ogilvie2017MinimizingTC,Huang2014ActiveLB}. Commonly,   
active learning approaches can be categorized based on the way they process items in the database: (i) Sequential Active Learning (SAL), in which one instance of data per algorithm iteration step is selected, and (ii) Batch-mode Active Learning (BAL), in which a batch of informative instances is processed. The shortcoming of SAL compared to BAL is that it requires re-training after each iteration, which can be quite costly \cite{Konyushkova2017LearningAL,Haumann2019DeepAL}. The main goal of BAL is to select a batch of unlabeled samples on each iteration, trading off between \emph{uncertainty} and \emph{representativeness} \cite{Huang2014ActiveLB}. Uncertain samples are likely helpful to refine the decision boundary, while representativeness (according to some surrogate criterion) strives to capture the data structure of the unlabeled set. There are two main active learning approaches to achieve this trade-off: (i) objective-driven methods and (ii) screening approaches. \emph{Objective-driven methods} formulate this trade-off in a single objective, such that the most informative/representative batch of samples is found by solving an optimization problem~\cite{Schohn2000LessIM, Hoi2009BatchMA}. These approaches are often theoretically well-grounded and have demonstrated good performance in practice. However, they usually do not scale well to large datasets \cite{Lourentzou2018ExploringTE}. Although these works show promising results, most of them are performed on a balanced dataset that does not simulate the imbalanced data distribution in real-world applications. \\
Early work on imbalanced active learning was done by Ertekin et al.~\cite{Ertekin2007LearningOT}. 
The proposed framework was able to solve the problem of class imbalance in active learning by providing more balanced samples to the active learning classifier. The study demonstrated an efficient technique to select informative samples from a smaller sample for active learning. The authors pointed out that implementing early stopping criteria enables active learning to achieve a fast solution with competitive performance for an imbalanced class distribution. While other works \cite{Lin2018ActiveLW,Aggarwal2020ActiveLF} prefer balancing samples after selecting the batch-mode active learning algorithm for imbalanced datasets. Bhattacharjee et al. \cite{ALOD} proposed a current active learning framework to address the unbalanced distribution of classes. The study proposed two algorithms that used a novel sampling strategy and an anomaly detection method. \\
Recently many works have studied active learning methods using deep neural networks and referred to as Deep Active Learning (DAL), and these works have been reviewed by \cite{Mohamadi2020DeepBA,Ren2022,Liu2022,Zhan2022ACS}. 
For example, Ren et al. \cite{Ren2022} conducted a comprehensive study on the potential of deep learning in the context of active learning. The study involved a study of various deep active learning strategies and architectures, as well as an assessment of their impact on overall performance, learning efficiency, and field application. The authors also explored several key topics related to deep active learning, including query strategy optimization (batch-mode DAL, uncertainty-based query and density-based query), labeled sample data expansion (cost-effective active learning and generative adversarial active learning), model generality (heuristic deep active learning), and stopping strategy. 
On the other hand, Liu et al. \cite{Liu2022} proposed to categorize deep active learning techniques into two primary types: model-driven and data-driven. Model-driven techniques make use of the internal states of the model, such as prediction confidence or feature representations, to select the most informative samples. In contrast, data-driven techniques rely more on the characteristics of the data itself, like the density or diversity of the samples, to guide the active learning process.
\subsection{Fair Machine Learning}
There are plenty of distortions that make it hard to process data fairly \cite{ChouldechovaR20}, including biased encoded data and the effect of minimizing average error to fits majority populations. 
Fair Machine Learning (FML) aims to present a machine learning model that is free of bias toward sensitive groups. Fairness in machine learning has attracted tremendous research interest over several years \cite{mehrabi2021survey, gajane2022survey}. It has been explored in numerous data mining and machine learning problems.
One of the key research on fairness was introduced by Dwork et al. \cite{Dwork2012FairnessTA}, in which they introduced two notions of fairness, (i) \emph{individual fairness} and (ii) \emph{group fairness}. Individual fairness is focused on consistent treatment and strives to achieve configurations where similar objects are assigned similar outcomes. On the other hand, group fairness ensures that results are equally distributed across all subgroups defined based on sensitive attributes, such as gender, race, ethnicity, nationality, and religion. Many research works have highlighted the bias of ML models for different sensitive attributes, and therefore it must be protected. \cite{Kleindessner2020ANO,Gupta2019IndividualFI}. 

Following these two directions, fair clustering has also been widely studied by the FML community. Individual fairness in clustering focuses on answering the question of how to treat similar individuals similarly. Some representative studies include similar samples receiving similar cluster placement~\cite{brubach2021fairness} and similar samples receiving similar cluster distance \cite{vakilian2022improved}. Meanwhile, group fairness in clustering focuses on treating each sample in the group fairly with respect to how other groups are being treated. One of the key ideas is to balance clustering representation in clustering algorithms, e.g., spectral clustering \cite{kleindessner19b} and correlation clustering \cite{ahmadian20a}. 
\subsection{Fair Active learning}
Recently, the active learning community has been interested in investigating fairness during active learning sample selection. The first fair active learning method is proposed by Anahideh et al. \cite{Anahideh2020FairAL}. This model is targeted to tackle fairness in the group setting. The study focused on selecting fair samples to be labeled. They designed an active learning algorithm to select data vectors to be labeled by balancing the model's accuracy and fairness. The fairness was measured as the expected fairness over all unlabeled data and the newly added labeled data. Their proposed FAL method used an accuracy-fairness optimizer to select samples to be labeled and used three strategies for the optimizer: FAL $\alpha$-aggregate, FAL Nested, and FAL Nested Append. The FAL method notably reduced unfairness while not significantly impacting the model accuracy, and the FAL Nested Append optimizer showed the best performance across different experiments and fairness models. 
In recent work, Sharaf et al. \cite{Sharaf2020PromotingFI} formulated the fairness-constrained active learning as a bi-level optimization problem targeted for solving group fairness. The first level of optimization focuses on training a classifier on a subset of labeled examples. In contrast, the second level is the selection policy to choose the subset of data that achieves the desired fairness and accuracy performance on the trained classifier. To achieve this goal, they used forward-backward splitting on both optimization methods. The study showed a promising result in terms of accuracy; however, the fairness performance was not able to outmatch the previous fairlearn \cite{Agarwal2018ARA} method. More recently, Shen et al \cite{Shen2022MetricFairAL} take another path of studying fair active learning. Their work investigates metric-fair active learning of homogeneous halfspaces, demonstrating that fairness and label efficiency can coexist under the distribution-dependent PAC learning model.

\section{Problem Formulation}\label{sec:problem}
In this paper, we focus on the pool-based active learning problem. In pool-based active learning, a small labeled dataset $\mathcal{L}=\{(x_1,y_1),...,(x_n,y_n)\}$, consisting of pairs of feature vectors $(x_i)$ and their associated labels $(y_i)$, is available. The main focus of active learning is to improve the performance of a classifier $\mathcal{G}$ on a large pool of unlabeled data $\mathcal{U}=\{(x_1,...,x_n)\}$, which includes feature vectors $(x_i)$ that have not yet been assigned labels. The labeled dataset $\mathcal{L}$ is created by a domain expert, while the labels for $\mathcal{U}$ remain unknown. The goal is to select the most informative samples from the unlabeled pool to be labeled by an oracle and added to the labeled dataset $\mathcal{L}$, which is then used to update the classifier $\mathcal{G}$. The active learning process involves three components: an oracle that provides labeled data, a predictor that learns from the labeled data, and an acquisition function (AF) that guides the oracle in selecting the most informative samples to label. The AF maps the unlabeled data in $\mathcal{U}$ to an ordered set $\mathcal{V}$, which prioritizes the samples based on their expected usefulness in improving the classifier's performance. Typically, an AF is defined as the combination of (1) a feature extraction function $f(x):X \to Y$, where $x \in X$ is a feature vector from a set of samples $\{x_1,...,x_n\}$, that maps a sample to its corresponding class label in $Y$, and (2) a score function $g(y):Y \to V$, where $y \in Y$ is a class label, that calculates the uncertainty score of a sample. The AF is represented by the formula:
$$ AF = \sum\limits_{i=1}^{n} g(f(x_{i})).$$
Acquisition functions can be implemented using various approaches, such as traditional machine learning models (e.g., SVM) or advanced neural networks. Each method has its own strengths and weaknesses. Traditional machine learning models are generally more efficient than neural networks, but neural networks tend to produce more accurate predictions.

In detail, each iteration involves utilizing a classifier model $\mathcal{G}$ trained on the labeled dataset $\mathcal{L}$ to identify the most informative sample from the entire unlabeled dataset $\mathcal{U}$ (usually those samples that are close to the classifier boundary and exhibit high uncertainty). The selected sample is then forwarded to an oracle, typically a human expert, for labeling to its corresponding class. The objective of a good active learning method is to achieve a reasonably accurate classifier while minimizing the amount of labeling required from the oracle. The design of Acquisition Functions ensures that their maximums are located at regions of the input space $\mathcal{U}$ with high levels of uncertainty and in areas that have not been adequately explored. Commonly, manual labeling of instances for the oracle is often done one at a time to update the classification model, which can be a time-consuming and inadequate process that cannot be parallelized. A potential solution to this issue is the Batch-mode Active Learning (BMAL) approach \cite{Chakraborty2015AdaptiveBM,Guo2007DiscriminativeBM, Hoi2009BatchMA}, which allows multiple query instances to be processed simultaneously by forming batches. However, this approach can still be resource-intensive, particularly in real-world labeling processes, especially when a human expert is involved. A good BMAL algorithm should address this issue by providing a parallel processing procedure. In BMAL, a batch $b$ of unlabeled items from the dataset $\mathcal{U}$ is selected based on an objective function, subject to the constraint that the total batch size $b$ cannot exceed the budget size $|B|$ and the batch size is determined by human labeling ability. The AL method selects an unlabeled sample from $\mathcal{U}$ for labeling by a domain expert in each iteration until a stopping criterion is reached, such as a model accuracy threshold or an exhausted labeling budget.

The issue of imbalanced unlabeled data raises the problem of selecting a batch that adequately represents all classes and ensures fairness in processing. Hence, there is a need to develop fairness-aware active learning techniques. The research in this field can explore various directions since there are different definitions of fairness. In this paper, we focus on statistical fairness, which is commonly considered fairness for active learning research. The objective of such a method is to choose representative samples for batch $b$ that maintain classifier performance while also ensuring fairness for specific sensitive groups.
Thus, we can frame our problem as a multi-objective optimization (MOP) as follows 
\begin{equation}
    MOP=max(A(X^*)) + min (UF(X^*))
\end{equation}
where $(X^*)$ is the set of selected samples derived from unlabeled data $\mathcal{U}$, and the first objective function $A$ is to maximize the performance of the model, e.g., the accuracy of the predictions, while the second objective function is to minimize the unfairness $(UF)$.  

In order to formulate the algorithm, it is important to first identify the sensitive attribute that will be used. In this paper, we choose race or gender as the sensitive attribute. For each unlabeled data point, denoted as $x \in \mathcal{U}$, we have a set of sensitive attributes defined as $S$. To represent the sensitive attribute of a specific data point $x_i \in \mathcal{U}$, we use the notation $S^i$.
\begin{figure*}
    \centering
    \includegraphics[width=4in]{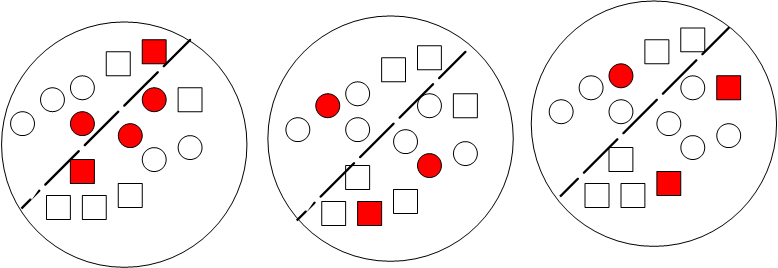}
    \caption{The FAL-CUR methods illustration compared to uncertainty and representativeness sampling method. The uncertainty sampling method selects samples that the classifier is uncertain about, while representativeness sampling picks the samples which are representative of all samples inside the cluster. The FAL-CUR uses a weighted score of uncertainty and representativeness.}
    \label{fig:illustration}
\end{figure*}

\section{The Proposed Method: FAL-CUR}\label{sec:methodology}
Since we face imbalanced and fairness constraints, it is essential to select the samples that can represent the minority class and has a higher fairness score. We propose a novel method, called \textbf{F}air \textbf{A}ctive \textbf{L}earning using fair \textbf{C}lustering \textbf{U}ncertainty and \textbf{R}epresentativeness (FAL-CUR). FAL-CUR combines fair clustering with uncertainty and representativeness score as the query strategy to achieve high fairness performance while maintaining the accuracy of the model.  
Fig \ref{fig:illustration} illustrates the sample selection of the proposed FAL-CUR method. The proposed FAL-CUR model is constructed into two main parts. In the first part, we cluster the uncertain samples using fairness aware clustering \cite{Abraham2020FairnessIC}. In the second part, the method focuses on selecting the representative samples that reside in different regions inside the fair cluster. The representative samples are considered as the ones with high uncertainty and representative of the whole sample. 

\subsection{Fair Clustering}
Our fair clustering method is inspired by FairKM (short for Fair KMeans)~\cite{Abraham2020FairnessIC}. FairKM extends KMeans clustering algorithm to further incorporate fairness towards sensitive attributes, e.g., gender and race. The objective function of FairKM consists of two parts: clustering and fairness. The clustering component, which is directly from KMeans, aims to minimize the distances between the data points and their corresponding cluster centroids. The fairness component aims to minimize the deviation of fairness on sensitive attributes. Formally, given the dataset $\mathcal{X}=\{...,X,...\}$, the set of non-sensitive attributes $\mathcal{N}$ and the set of sensitive attributes $\mathcal{S}$, the objective function is defined as:
\begin{equation}
\mathcal{O}=\sum_{C \in \mathcal{C}}\sum_{X \in C} dist_\mathcal{N}(X,C) + \lambda \: fairness_\mathcal{S}(C,\mathcal{X})
\label{eq:fairclustering_term}
\end{equation}
where $C$ indicates one cluster, $dist_\mathcal{N}(X,C)$ represents the standard KMeans loss, $fairness_\mathcal{S}(C,\mathcal{X})$ measures the fairness deviation, and $\lambda$ controls the balance of clustering and fairness performance. Since the fairness loss term is defined over attributes in the sensitive attribute set $\mathcal{S}$, it is required to define the deviation measure over individual attributes. Given a cluster $C$, a specific attribute $S$, and a choice of value $s$ for $S$ which is represented as $X.S=s$, the deviation between the fractional representations in $\mathcal{C}$ and $\mathcal{X}$ is defined as:
\begin{equation*}
    D_{C}^{S}=\left( \frac{|\{X|X \in C \wedge X.S=s\}|}{|C|} 
    -\frac{|\{X|X\ \in \mathcal{X} \wedge X.S=s\}|}{|\mathcal{X}|}
    \right)^2 
\end{equation*}\label{eq:2}
Then for all possible values of $S$ the deviation is summed over all possible values and normalized by the number of different values that attribute $S$ can take. This results in a normalized attribute-specific deviation $ND_{C}^{S}$:
\begin{equation*}
ND_{C}^{S}=\frac{D_{C}^{S}}{|Values(S)|}
\end{equation*}
where $|Values(S)|$ is the number of different values that attribute $S$ can take. Considering all attributes in $ \mathcal{S}$, the total deviation is defined as the sum of all sensitive attributes for the cluster $C$:
\begin{equation*}
ND_C=\sum_{S \in \mathcal{S}}D_{C}^{S}
\end{equation*}
To ensure equal treatment of all clusters, this total fairness deviation is weighted by the size of each cluster. \begin{equation*}
\sum_{C \in \mathcal{C}} \left(\frac{|C|}{|\mathcal{X}} \right)^{2} \times ND_C
\end{equation*}
Finally, the overall fairness deviation of fair clustering can be calculated as the deviation of the sensitive attribute in clusters and the whole dataset. 
\begin{equation}
    fairness_\mathcal{S}(\mathcal{C},\mathcal{X})=\sum_{C\in \mathcal{C}} \left(\frac{|C|}{|\mathcal{X}} \right)^{2} \times \sum_{S\in \mathcal{S}} \frac{\sum_{s\in Values(s)}\left(Fr_{C}^S(s)-Fr_{\mathcal{X}}^{S}(s) \right)^2}{|Values(S)|}
    \label{eq:cluster_fairness_all}
\end{equation}
where $Fr_{C}^S(s)$ and $Fr_{\mathcal{X}}^{S}(s)$ are shorthand for the fractional representations of the objects that take the value $s$ of the attribute $S$ (i.e., $S=s$) in $C$ and $\mathcal{X}$ respectively. The smaller the deviation value is, the fairer the clustering results we can obtain because smaller $fairness_\mathcal{S}(\mathcal{C},\mathcal{X})$ indicates a smaller deviation between the proportions of sensitive attributes in the clusters and the entire dataset.

Our work, however, aims to calculate the fairness deviation for each cluster in order to rank clusters. Therefore, we extend equation \ref{eq:cluster_fairness_all} by removing the sum over all clusters, which results in a cluster-specific fairness deviation measure:
\begin{equation}
F_{fairness}(C,\mathcal{X}) = \bigg( \frac{|C|}{|\mathcal{X}|}\bigg)^2 \times \sum_{i \in S} \frac{\sum_{t \in x_{i}}(Fr_C^S(t)-Fr_{\mathcal{X}}^S(t))^2}{|x_{i}|}
\label{eq:cluster_fairness}
\end{equation}
Finally, all clusters are ranked from most fair to most unfair using the cluster-specific fairness deviation scores, i.e., equation \ref{eq:cluster_fairness}.

\subsection{Sample Selection}
Since we have provided fairness through fair clustering, the next challenge is to select the candidate sample from the unlabeled pool to be labeled, improving the model with a limited budget. Thus, it is essential to search for the best representative sample inside the fair cluster. In this part, we consider a distance-based candidate selection that meets several criteria, such as having a high probability of belonging to a minority class and reducing misclassification error or model uncertainty. The candidate we are searching for could reside in several locations, such as near the centroid of the fair cluster or far from the fair cluster centroid. Thus the objective function can be denoted as follows.
\begin{equation}
    X^*=\underset{x\in \mathcal{U}}{\argmax} \; \mathcal{H}_t(y|x,\mathcal{G})+ \; d(x^i,C)
\end{equation}
\paragraph{\textbf{Uncertainty Score}}
We consider the uncertainty as the entropy of the classifier for a specific sample and its sensitive attribute. $\mathcal{H}_t(y|x,c)$ is the entropy of specific sample $(x_i,s_i) \in \mathcal{U}$ with regards to its classifier $\mathcal{G}$ trained in $t$-time. While $\mathcal{Y}$ is all the possible class values. The entropy can be calculated as follows
\begin{equation}
   \mathcal{H}_t(x) = -\sum_{y \in \mathcal{Y}} P(y | x, \mathcal{G}) \log P(y | x, \mathcal{G}),
   \label{eq:entropy}
\end{equation}

\paragraph{\textbf{Representative Score}}
Although we reduce the sample searching space by implementing fair clustering, it is still a question that which samples should be selected for labeling. In this work, we consider using representative and uncertainty sampling for sample selection. After measuring the uncertainty, the next step is to calculate the representative score of each sample. For this, we calculate the function of representativeness $Rep$, which calculates the most similar sample to all the samples using Euclidean distance.
Consider $x_s^{i}, x \in \mathcal{U}$ is the sample grouped by fair clustering and the ground-truth class label of $i$-th sample inside the fair cluster $x_s^{i}$ is unknown, we define representativeness of $x_s^{i}$ if $x_s^{i}$ share a large similarity $-$ measured by euclidean distance $(d)$ $-$ with other samples inside the cluster. Thus, we measure the representative score by calculating a pairwise comparison for each sample: 
$$ Rep(x) = \sum\limits_{i=0}^{n} \sum\limits_{j=0}^{n} d(x_s^{i},x_s^{j}).$$

\paragraph{FAL-CUR Sample Acquisition}
The final score for sample selection can be defined as follows:
\begin{equation}
x^*= \beta Rep(x_s^{i})+(\beta-1)\mathcal{H}(x_s^{i}) \label{eq:sample_selection}
\end{equation}
We rank this score and select the samples with the highest score. 
The $\beta$ parameter is responsible for weighing the impact of the two scores on the final score. Small $\beta$ makes the model favors uncertainty, while higher means the model prefers representatives. In our experiments, we fixed $\beta$ to 0.6, which was the balance of the two scores. 

\begin{figure*}
    \centering
    \includegraphics[width=\textwidth]{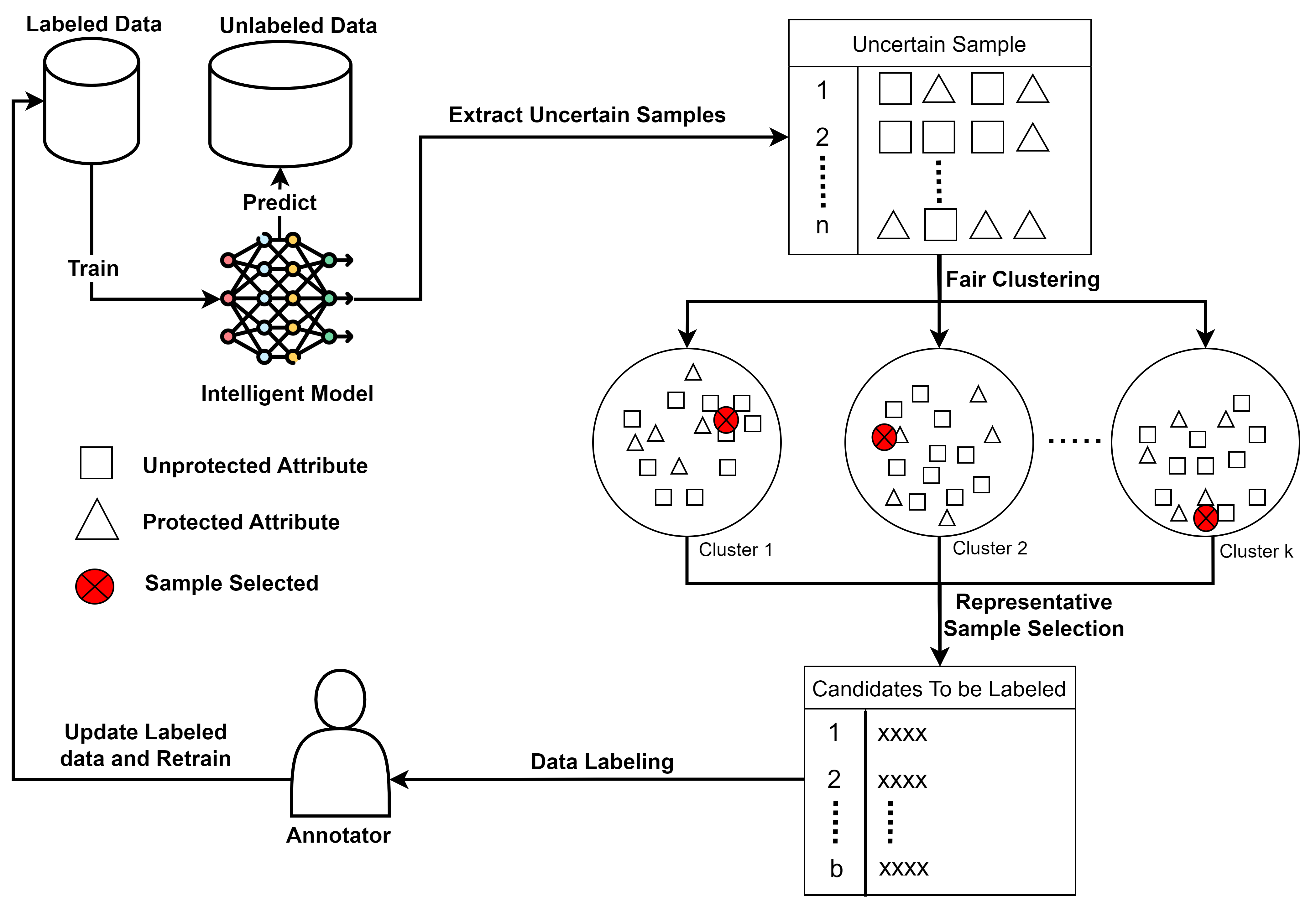}
    \caption{The flowchart of FAL-CUR method. The fair clustering approach groups samples together based on the proportions of protected attribute groups within each cluster. This reflection of the dataset ensures fairness in the clustering process. 
    } 
    \label{fig:flowchart}
\end{figure*}

Figure \ref{fig:flowchart} illustrates the flowchart for the proposed FAL-CUR method. Essentially, we train a machine learning model such as logistic regression on labeled data and use it to predict the uncertainty of samples in an unlabeled data pool. The extracted samples will be then clustered using the fair clustering method \cite{Abraham2020FairnessIC}. Next, the fair clusters would be ranked based on their fairness scores. Then we iteratively select the highest ranked samples from ranked cluster until $b$ samples are selected for a batch. The proposed method employs a batch size parameter $b$ that might not always be equal to the number of clusters \textit{k}. When the batch size surpasses the number of clusters, we restart the sample selection process, starting again by selecting the highest-ranked sample from the topmost cluster. Conversely, if the batch size is less than \textit{k}, our selection is confined to samples solely from the top-ranked \textit{k} clusters. In simple words, first we select the first ranked sample from the highest to the lowest ranked cluster, then second ranked sample in the same manner, and so on. The process is stopped when $b$ samples are selected. 
Finally, the selected samples are sent to human experts to be labeled. The newly labeled data will be added to the labeled data pool and will be used to retrain the model.

The proposed FAL-CUR method is explained in detail in Algorithm \ref{alg:cap}, and all variables that will be used are explained in Table \ref{tab:variables}.

\begin{algorithm*}[htb!]
\caption{FAL-CUR}\label{alg:cap}
\SetAlgoLined
\KwIn{Labeled Dataset $\mathcal{L} $ $={(x_1,y_1),...,(x_n,y_n)}$; 
Unlabeled Dataset $\mathcal{U}={(x_1,...,x_m)}$;
Batch size $b$; Number of clusters $k$;
Number of iterations $T$; Classifier $\mathcal{G}$;
}
\KwOut{Selected samples $X^* ={(x_1^*,...,x_b^*) \in \mathcal{U}}$ }
\For{$t=1,2,...,T$}{
Identify $k$ clusters using fair-clustering method;\newline
$\tau_C$= list of ranked clusters based on their fairness score;\newline
\For{each sample $x \in C$}{
    Calculate sample score of $x \in C$ using equation \ref{eq:sample_selection}\newline
    Rank samples within the cluster based on their scores
}
$selected\_samples=0$; \newline
j=1; \newline
\While{$selected\_samples < b$}{ 
\For{$i=1,2,...,k$}{
\If{$selected\_samples < b$}{
Select $j_{th}$ ranked sample from $\tau_C[i]$ and add it to the candidate sample set $X^{*}$;\newline
$selected\_samples = selected\_samples +1$;
}
$j=j+1$;
}
}
Obtain labels $(x^{*},y)$ for the samples in $X^{*}$, add them to $\mathcal{L}$, and remove $X^{*}$ from $\mathcal{U}$;\newline
Update the classifier $\mathcal{G}$ using the newly labeled samples in $\mathcal{L}$;
}
\end{algorithm*}

\begin{table}[htbp]
\caption{Variables in Algorithm 1}
\centering
\scalebox{0.70}{
\begin{tabular}{c|l}
\hline \hline
Variable & Explanation \\ \hline \hline
 $\mathcal{L}$        & Labeled dataset consisting of $n$ samples $\{(x_1,y_1),...,(x_n,y_n)\}$.            \\ \hline
 $\mathcal{U}$        & Unlabeled dataset consisting of $m$ samples $\{(x_1,...,x_n)\}$.\\ \hline
 $T$        &  Number iteration \\ \hline
 $\mathcal{G}$        &  Classifier used for training on the labeled dataset.           \\ \hline
  $C$        &  Set of clusters.           \\ \hline
 $k$        &    Number of clusters used in the clustering algorithm.         \\ \hline
 $\tau_C$ & list of clusters ranked based on their fairness score \\ \hline
 $\tau_C[i]$ & $i_{th}$ ranked cluster based on their fairness score \\ \hline
  $S$        &   Set of sensitive attributes appear in the dataset.          \\ \hline 
 $X^*$        &   Sample selection, i.e., the set of samples to be labeled in the next iteration.           \\ \hline \hline
\end{tabular}}
\label{tab:variables}
\end{table}

\subsection{Computational Complexity Analysis}
The computational complexity of FAL-CUR can be analyzed by examining its individual components and taking the most complex process. 
\begin{itemize}
    \item \textbf{Identify k clusters using fair-clustering method.}\\ Since fair clustering uses KMeans thus the complexity would be $\mathcal{O}(nTkI)$, where $n$ is the number of samples, $T$ is the number of iterations of the algorithm, $k$ is the number of clusters, and $I$ is the average number of iterations needed for convergence in the KMeans algorithm.
    \item \textbf{Ranking clusters based on their fairness score}\\ This would require calculating the fairness score for each cluster and then sorting them. This step would take $\mathcal{O}(k\log k)$.
    \item \textbf{Ranking samples within each cluster}. \\ For each of the fair clusters, the algorithm needs to calculate a sample score and rank the samples within that cluster. Thus, for one cluster, assume the number of samples is $m$, the complexity of this step is $\mathcal{O}(m \log m)$.  
   \item \textbf{Sample selection.} \\ The algorithm might need to iterate over all the samples in all the clusters. This would be $\mathcal{O}(n)$.
\end{itemize}
Thus, the overall time complexity of the FAL-CUR algorithm would be dominated by two steps i.e., fair clustering and sample selection. The fair clustering step takes the complexity of $\mathcal{O}(nTkI)$ while sample selection and sample ranking take $\mathcal{O}(Tkm \log m)$. 

\section{Experimental Setup}\label{sec:expsetup}
In this section, we discuss the experimental setup for our analysis, including datasets, evaluation methods, and baselines.
\subsection{Datasets}
\begin{table*}[htbp]
\centering
\caption{An Overview of the datasets}
\scalebox{1}{
\begin{tabular}{p{1.1cm}|C{1.6cm}|C{1.7cm}|C{1.5cm}|C{1.5cm}|C{1.5cm}}
\hline \hline
Dataset & \#Instances & \#Attributes & Positive Class & \%Minority Class & Sensitive Attribute
\\ \hline 
Adult                    & 32560                        & 15                            & \textgreater{}50K               & 24\%                               & Gender                               \\ \hline
Compas                  & 4483                         & 9                             & Recidivism                      & 26\%                               & Race                                 \\ \hline
Loan                     & 5000                         & 16                            & Verified                        & 34\%                               & Gender                               \\ \hline 
OULAD               & 21562                       & 12                            & Pass               & 32\%                                & Gender                               \\ \hline
\hline
\end{tabular}}\label{tab:Datasets}
\end{table*}

We use four real-world datasets for the analysis, and the characteristics of these datasets are presented in table \ref{tab:Datasets}. The datasets vary in dimensionality and class imbalance, and therefore they are interesting for comparative evaluation. Next, we explain them in detail. 

\subsubsection{Adult census income \cite{kohavi1996scaling}} 
This dataset contains 32,560 records extracted from 1994 census data. The attribute for the dataset includes age, occupation, education, and gender. The class is defined based on income such that it is 1 if the income is higher than \$50k and 0 otherwise. The sensitive attribute is gender, with $s=Female$ being the protected group. 
\subsubsection{Compas dataset \cite{angwin2016machine}} 
The Compas ProPublica dataset\footnote{https://www.propublica.org/datastore/dataset/compas\-recidivism\-risk\-score\-data\-and\-analysis} includes attributes describing the sex, age, race, juvenile felony, and misdemeanor counts, number of adult priors, charge degree (felony or misdemeanor) in the state of Florida USA. We filtered the dataset to divide the race (protected attribute) into White and African-American defendants, as discussed in \cite{Anahideh2020FairAL}. 
The final dataset consists of 4483 records, and the two-year recidivism feature is used as the output label. In the Compas dataset, 26\% of the data belongs to the minority class.

\subsubsection{Loan application data} 
This dataset is extracted from loan application data and is available at \url{https://github.com/h2oai/app-consumer-loan}. We preprocess that dataset, including removing null values and scaling, and the final cleaned dataset has a total of 5000 records collected from 2007 to 2015. The main features include the loan amount, purposes, and gender. The gender feature is used for defining a protected group. The task is to determine whether a loan application has a good or bad risk. 
\subsubsection{OULAD \cite{kuzilek2017open}} 
The Open University Learning Analytics (OULAD) dataset contains information about students and their activities in the virtual learning environment (VLE) for 7 courses. The dataset contains information of 32,593 students characterized by 12 attributes (7 categorical, 2 binary, and 3 numerical attributes). The target of the prediction task on the class label is the final result of the student = {pass, fail}. We use the cleaned dataset with 21,562 instances after removing the missing values and rows with the final result = ``withdrawn".
Gender is used as the protected attribute, and the ratio male:female is 11,568:9,994 (56.6\%:46.4\%).

\textcolor{black}{\subsection{Preprocessing Steps}
The dataset used in this study has several challenges such as the presence of missing values, categorical variables that needed to be represented in a numeric format, and numerical data with differing scales that needs to be normalized. 
The existence of these challenges makes it difficult to make use of most machine learning models to analyze the data.
The data preprocessing was carried out using the following three steps. 
\begin{itemize}
    \item \textbf{Missing Value Handling}: 
    There were some occurrences of missing values in the dataset - usually represented as `Null' or `NaN'. Unfortunately, most predictive modeling methods were unable to process these missing values effectively \cite{kuhn2019feature}. Thus, our initial step in preprocessing was to identify and eliminate these missing values from the dataset. 
    \item \textbf{Categorical Variable Conversion}: We converted categorical variables into a numerical format and used one-hot encoding to achieve this transformation. 
    \item  \textbf{Min-Max Normalization}: Finally, we normalized values using min-max normalization to re-scale the numerical variables in a range between 0 and 1. 
    \end{itemize}
}

\subsection{Evaluation Criteria}
The evaluation of the proposed method is based on two distinct criteria: Performance of the model as well as its fairness.
\subsubsection{Performance Measures} 
Three metrics are used to assess model performance using the confusion matrix as described in Table~\ref{tab:confusion_matrix}.

\begin{table}[htbp]
\centering
\caption{Confusion Matrix}
\begin{tabular}{cc|cc}
\hline \hline
\multicolumn{2}{c|}{\multirow{2}{*}{}} & \multicolumn{2}{c}{Actual Class} \\ \cline{3-4} 
\multicolumn{2}{c|}{}                  & Positive (P) & Negative (N) \\ \hline
\multirow{2}{*}{\rotatebox[origin=c]{0}{Predicted Class}} & Positive (P) & True Positive (TP)  & False Positive (FP) \\ \cline{2-4} 
 & Negative (N) & False Negative (FN) & True Negative (TN) \\  \hline \hline
\end{tabular}\label{tab:confusion_matrix}
\end{table}

All these measures are defined below. 
\begin{itemize}
    \item Accuracy: It determines the fraction of all instances that are correctly classified using a machine learning model. The formula to compute accuracy is:
    \begin{equation}
    \text{Accuracy} = \frac{\text{TP+TN}}{\text{TP+FP+FN+TN}}
    \end{equation}
    \\While accuracy is commonly used in machine learning, it can be misleading especially in imbalanced data distribution. The accuracy could be high even if the model is biased towards predicting the majority class. 
    \item GMeans: This metric evaluates the balance between sensitivity (true positive rate) and specificity (true negative rate). It's particularly useful for imbalanced datasets. The GMeans formula is:
    \begin{equation}
    GMeans=\sqrt{\frac{TP}{TP+FN} \times \frac{TN}{TN+FP}} 
    \end{equation}
    \item F1-score: It is computed as: 
    \begin{equation}
    F1 =\frac{2 \times Precision \times Recall}{Precision+Recall}
    \end{equation}
where, Precision= $\frac{TP}{(TP+FP)}$ and Recall = $\frac{TP}{(TP+FN)}$. F1-score ranges between 0 and 1, where a value of 1 indicates perfect precision and recall, while a value of 0 indicates poor performance. The F1-score considers both precision and recall, making it a better indicator of overall performance.
\end{itemize}

\subsubsection{Fairness Evaluation} 
The methods are compared based on three fairness metrics which are defined below.
\begin{itemize}

    \item 
    Equal Opportunity Difference: This metric calculates the disparity in positive predictions between the protected ($S=1$) and non-protected ($S=0$) groups when the actual outcome ($Y$) is positive. The formula is:
    \begin{equation*}
    |Pr \{\hat{Y}=1|S=0,Y=1\}-Pr \{\hat{Y}=1|S=1,Y=1\}|
    \end{equation*}
    \item Average Equalized Odds Difference: This captures the average differences in both false positive rates and true positive rates between protected and non-protected groups. It is computed as: 
    \begin{equation*}
    \begin{aligned}
    | \frac{1}{2}[Pr\{\hat{Y}=1|S=0,Y=0\}-Pr\{\hat{Y}=1,S=1,Y=0\}]+ \\      \frac{1}{2}[Pr\{\hat{Y}=1|S=0,Y=1\}-Pr\{\hat{Y}=1,|S=1,Y=1\}] |
    \end{aligned}
    \end{equation*}
    
    \item Statistical Parity Difference: This measures whether each group (protected and non-protected) is predicted as positive at equal rates. The formula is:
    \begin{equation*}
        |Pr\{\hat{Y}=1|S=0\}-Pr\{\hat{Y}=1|S=1\}|
    \end{equation*}
\end{itemize}

For all fairness metrics, a value closer to zero shows that there is a small difference between protected and non-protected groups, which indicates a fairer prediction. Negative values indicate that the model is biased against the unprivileged group, while positive values indicate otherwise.

\color{black}
\subsection{Baselines Methods}
We compared our method with the following active learning and fair active learning methods.

\begin{itemize}
    \item {\bf{AL}}\cite{Lewis1994ASA}. We use a standard active learning algorithm that selects the most uncertain sample through entropy as the candidate to be labeled by a domain expert.
     \item {\bf{BW }} \cite{Aggarwal2020ActiveLF}. The BW (Balance Weighting) method consists of two parts. First is the acquisition function to select samples using a pre-trained neural network in the source domain. Secondly, a balancing step is added to the acquisition function to reduce the imbalance of the labeled subset. 
     \item {\bf{ALOD-RE}} \cite{ALOD}. ALOD-RE (Active Learning with Outlier Detection Techniques and Resampling Methods) was proposed for active learning in imbalanced domains. This model combines active learning with outlier detection to select the most suitable samples to be labeled. ALOD-RE divides the query strategy by selecting 70\% of the sample candidate by either uncertainty sampling or query by committee, while the rest of 30\% is selected by outlier detection. It also adds a resampling method to balance the sample candidate before labeling. 
    \item {\bf{EADA }} \cite{Xie2021ActiveLF}. The EADA (Energy-based Active Domain Adaptation) is a recent active learning approach that queries groups of target data that incorporate both domain characteristics and instance uncertainty into every selection round.     
    \item {\bf{FAL}} \cite{Anahideh2020FairAL}. Fair Active Learning (FAL) is the first approach that takes fairness into account for active learning. FAL focuses on developing active learning for solving group fairness, specifically improving demographic parity scores. 
    The FAL consists of two main components: the accuracy optimizer that selects the samples that are expected to reduce the miss-classification error and a decision approach that approximates the samples' expected unfairness reduction, i.e., the point that is expected to impart the largest reduction of the current model unfairness after acquiring its label.
        
\end{itemize}
Table \ref{tab:baselines_performance} compares the research problem focused by various baseline methods including the performance and fairness evaluation metrics considered by them. In the last row, we mention our proposed method and the evaluation metrics that we consider in this work. 

\begin{table}[htb]
\caption{The comparison of all baseline methods}\label{tab:baselines_performance}
\centering
\scalebox{0.7}{
\begin{tabular}{c|c|c|c}
\hline \hline
\textbf{Method} & \textbf{Focused Problem}  & \textbf{Metrics}  & \textbf{Fairness Metrics}  \\ \hline
AL & General Active Learning & E-Measure &  -  \\ \hline 
BW & Imbalanced Classification & Accuracy  & - \\ \hline 
ALOD-RE & Imbalanced Classification & GMeans & - \\ \hline 
EADA & Domain Adaptation & Mean Accuracy & - \\ \hline 
FAL & Fair Active Learning & Accuracy &  Statistical Parity  \\ \hline 
FAL-CUR & Fair Active Learning & Accuracy, F1-Score, GMeans & Statistical Parity, Equalized  \\
 &  & &  Odds, Equal Opportunity  \\
\hline \hline
\end{tabular}}
\end{table}

\subsection{Implementation Details}
A standard data cleaning process, including removing null values and normalization, is used to preprocess the datasets. 
For a fair comparison to the previous method~\cite{Anahideh2020FairAL}, we use logistic regression as the classifier in the implementation. \textcolor{black}{
In FAL-CUR logistic regression model, we employed a grid search method to optimize hyperparameters 'C' and `max\_iter' , while keeping the penalty type as `l2' and the solver as `liblinear'. The final values were selected based on the highest average score in a 5-fold cross-validation process.}
Next, we divide the dataset into three disjoint sets, such as 10\% training, 20\% test, and 70\% unlabeled data. We conduct experiments on all the datasets using all baselines, and the final score is calculated by taking the average over ten runs. We take 180 as the batch size for the sample selection as a real-lab experiment showed that humans were able to do 180 labels per hour \cite{Fajri2020PS3}. All experimental results are computed using this batch size. 
The code for our implementation with other experimental material is publicly available on Github\footnote{https://github.com/rickymaulanafajri/FAL-CUR}. 

\section{Results}\label{sec:results}
In this section, we discuss experimental results to demonstrate the fairness and performance of the proposed method as compared to baselines. We also study the impact of each component of the proposed method on its overall performance.

\begin{figure*}[!htb]
\centering
\subfloat[Adult - Statistical Parity ]{\includegraphics[width=.30\textwidth]{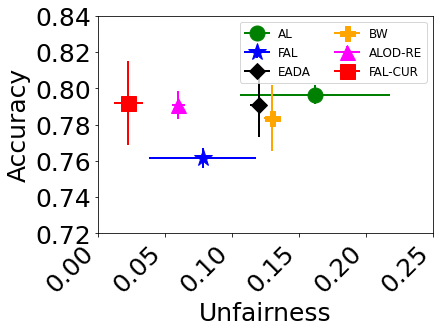}} \hspace{0.03\textwidth}
\subfloat[Adult - Equalized Odds ]{\includegraphics[width=.30\textwidth]{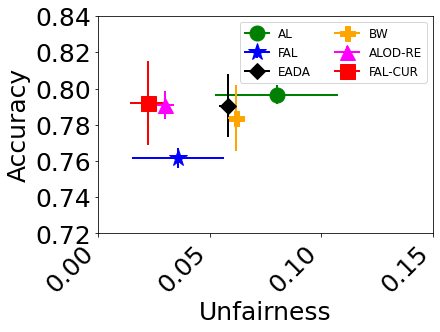}} \hspace{0.03\textwidth}
\subfloat[Adult - Equal Opportunity]{\includegraphics[width=.30\textwidth]{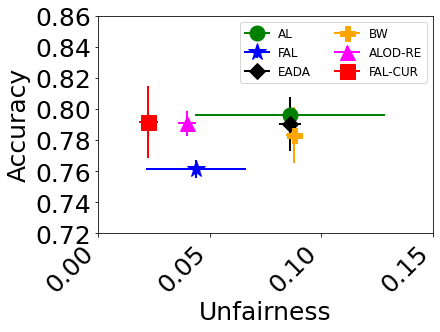}} \hspace{0.03\textwidth} \\
\subfloat[Compass - Statistical Parity]{\includegraphics[width=.30\textwidth]{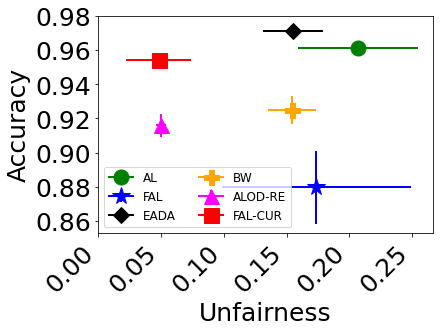}} \hspace{0.03\textwidth}
\subfloat[Compass - Equalized Odds]{\includegraphics[width=.30\textwidth]{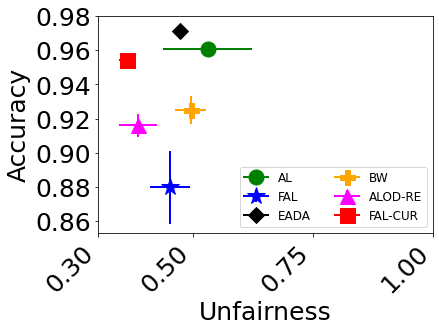}} \hspace{0.03\textwidth}
\subfloat[Compass - Equal Opportunity]{\includegraphics[width=.30\textwidth]{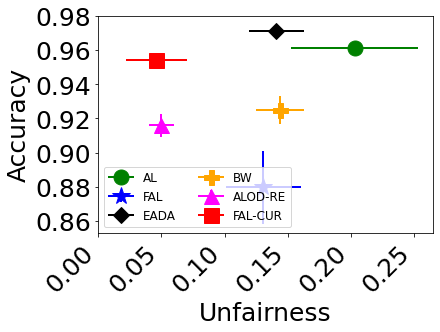}} \hspace{0.03\textwidth} \\
\subfloat[Loan - Statistical Parity]{\includegraphics[width=.30\textwidth]{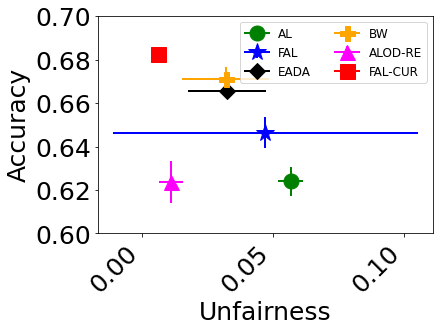}} \hspace{0.03\textwidth}
\subfloat[Loan - Equalized Odds]{\includegraphics[width=.30\textwidth]{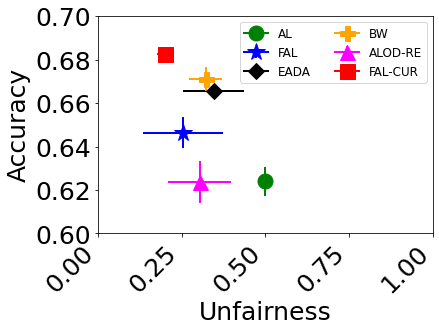}} \hspace{0.03\textwidth}
\subfloat[Loan - Equal Opportunity]{\includegraphics[width=.30\textwidth]{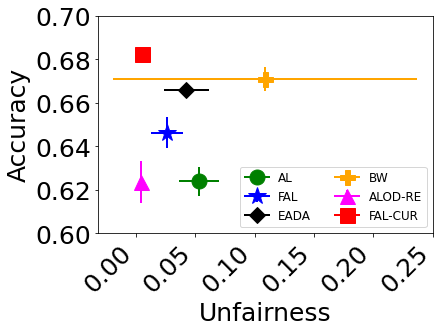}} \hspace{0.03\textwidth} \\
\subfloat[OULAD - Statistical Parity]{\includegraphics[width=.30\textwidth]{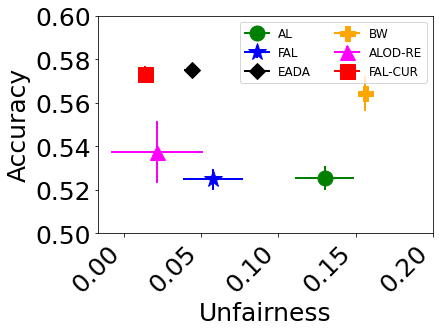}} \hspace{0.03\textwidth}
\subfloat[OULAD - Equalized Odds]{\includegraphics[width=.30\textwidth]{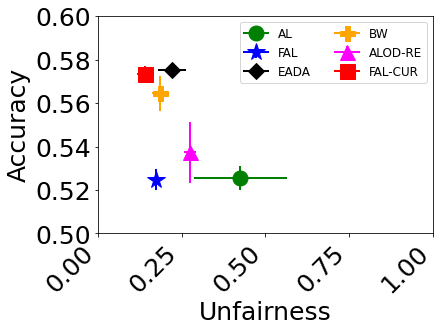}} \hspace{0.03\textwidth}
\subfloat[OULAD - Equal Opportunity]{\includegraphics[width=.30\textwidth]{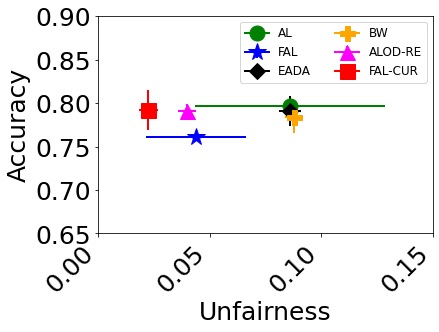}} \hspace{0.03\textwidth} \\
\caption{Fairness comparison of the FAL-CUR method with baseline methods on all datasets (Fairness Metrics against model accuracy).}\label{fig:Fairness Evaluation}
\end{figure*}

\begin{figure*}[!htb]
\centering
\subfloat[Adult - Accuracy ]{\includegraphics[width=.30\textwidth]{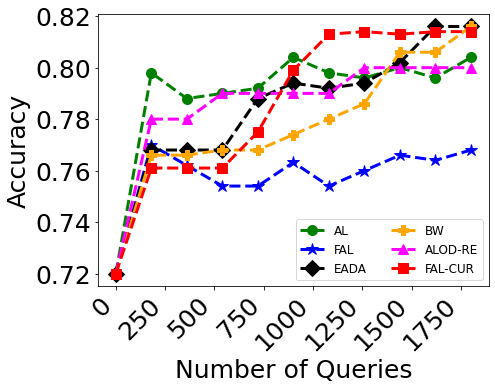}} \hspace{0.03\textwidth}
\subfloat[Adult - F1 Score]{\includegraphics[width =.30\textwidth]{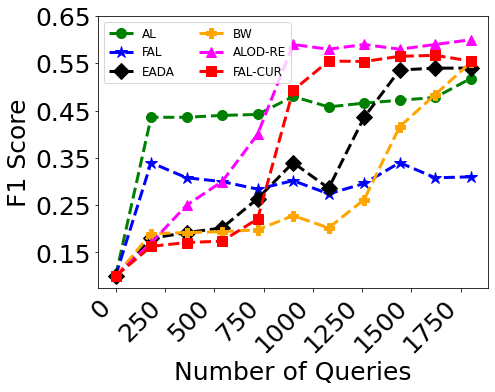}} \hspace{0.03\textwidth}
\subfloat[Adult - GMeans]{\includegraphics[width = .30\textwidth]{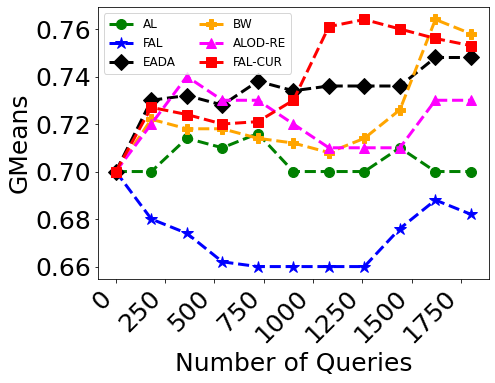}}\hspace{0.03\textwidth} \\
\subfloat[Compass - Accuracy]{\includegraphics[width = .30\textwidth]{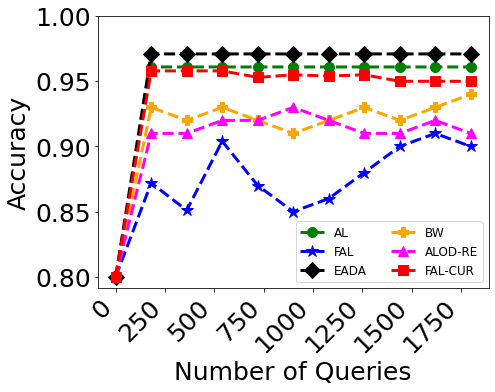}} \hspace{0.03\textwidth}
\subfloat[Compass - F1 Score]{\includegraphics[width = .30\textwidth]{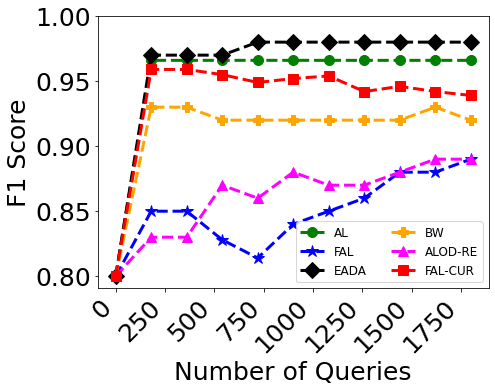}}\hspace{0.03\textwidth}
\subfloat[Compass - GMeans]{\includegraphics[width = .30\textwidth]{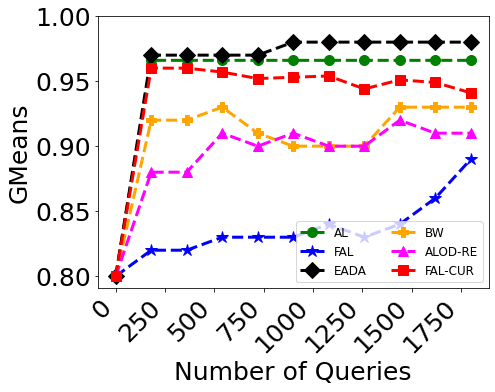}}\hspace{0.03\textwidth} \\
\subfloat[Loan - Accuracy]{\includegraphics[width =.30\textwidth]{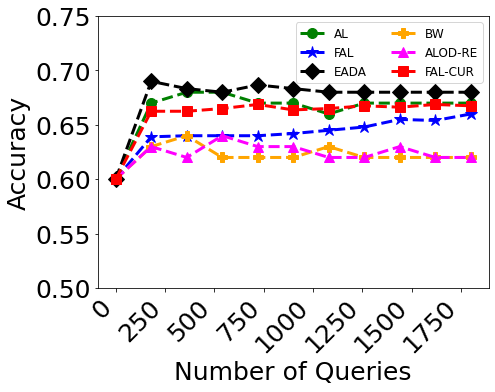}} \hspace{0.03\textwidth}
\subfloat[Loan - F1 Score]{\includegraphics[width = .30\textwidth]{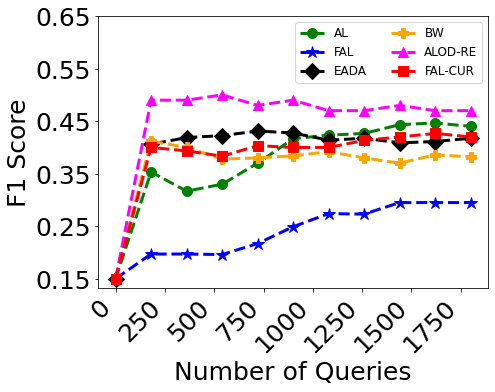}} \hspace{0.03\textwidth}
\subfloat[Loan - GMeans]{\includegraphics[width = .30\textwidth]{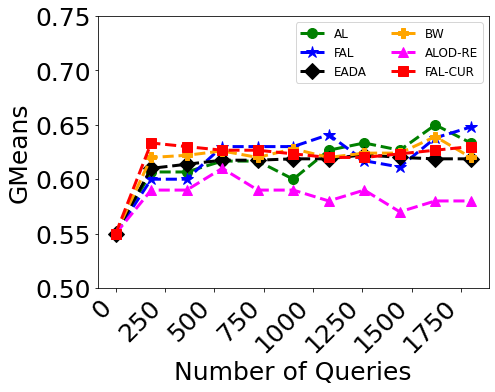}}\hspace{0.03\textwidth} \\
\subfloat[OULAD - Accuracy]{\includegraphics[width =.30\textwidth]{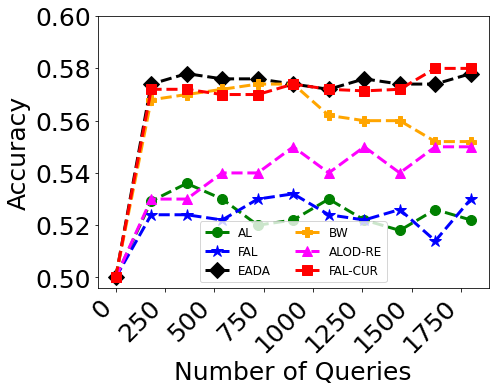}} \hspace{0.03\textwidth}
\subfloat[OULAD - F1 Score]{\includegraphics[width = .30\textwidth]{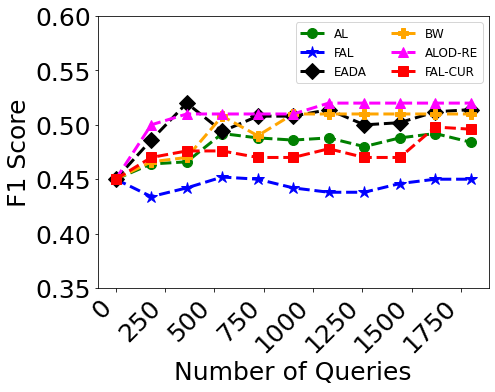}}\hspace{0.03\textwidth}
\subfloat[OULAD - GMeans]{\includegraphics[width = .30\textwidth]{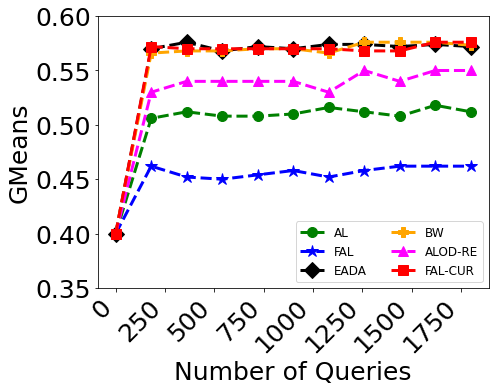}}\hspace{0.03\textwidth}
\caption{Comparison of the performance of FAL-CUR with baseline methods on multiple datasets (Metrics against the number of queries).}\label{fig:Performance Comparison}
\end{figure*}

\subsection{Fairness Evaluation}
We first evaluate fairness metrics, i.e., Statistical Parity (SP), Equal Opportunity (E\_Opp), and Equalized Odds (E\_Odd), for different methods, and the results are shown in Figures \ref{fig:Fairness Evaluation}. We plot models' performance using accuracy versus unfairness to measure how fairness is correlated with a model's performance in its outcome. The unfairness value towards 0 shows that the model is fairer. The proposed FAL-CUR method provides a better trade-off between the model's performance and fairness. In most cases (especially for Adult, Loan, and OULAD datasets), FAL-CUR provides better accuracy and fairness than the state-of-the-art fair active learning methods. ALOD-RE which is the latest fair active learning method performs second-best in terms of fairness, though it provides a much lower accuracy. FAL comes as the next fair model, followed by AL and EADA. On Adult and Compass datasets, the performance of state-of-the-art AL methods, such as EADA and BW, seems fairly close to each other; however, in almost all the cases, EADA provides the highest accuracy among baselines. The FAL-CUR method outperforms EADA in terms of all fairness metrics on all the datasets as well as in terms of accuracy on some of the datasets. 
We believe the key explanation for this result is that EADA focuses on selecting samples that are predicted to increase active learning performance. However, the samples chosen may be biased towards specific protected groups thus reducing its fairness score.
Furthermore, the result indicates that the proposed model works well and is fair in its labeling decision, which is supported by the fair sampling acquisition phase.  
\subsection{Performance Evaluation}
In this experiment, we compare how the performance of the FAL-CUR method behaves with the number of queries as compared to all baseline models. The results are shown in Figure \ref{fig:Performance Comparison}. The previous works \cite{Anahideh2020FairAL} showed that the model performance drops when fairness is maintained, which we also observed in our experiments. Figure \ref{fig:Performance Comparison}, clearly shows that methods not prioritizing fairness often yield superior results, albeit at the expense of lower fairness metrics. Therefore, in the proposed FAL-CUR method, we aim to improve fairness without largely harming the model performance. From the figure, we can summarize that EADA shows superior performance as compared to other models. However, its fairness score is lower as compared to the other models. The proposed FAL-CUR method is able to maintain high fairness performance without losing much in the classification performance. For example, on the Compass dataset, the accuracy of FAL-CUR is only 0.02 lower than the EADA and traditional AL methods. However, the fairness performance of FAL-CUR is the highest as compared to state-of-the-art models. 

\subsection{Ablation Study}
\begin{figure*}[!htb]
\centering
\subfloat[Performance]{\includegraphics[width = 0.4\linewidth]{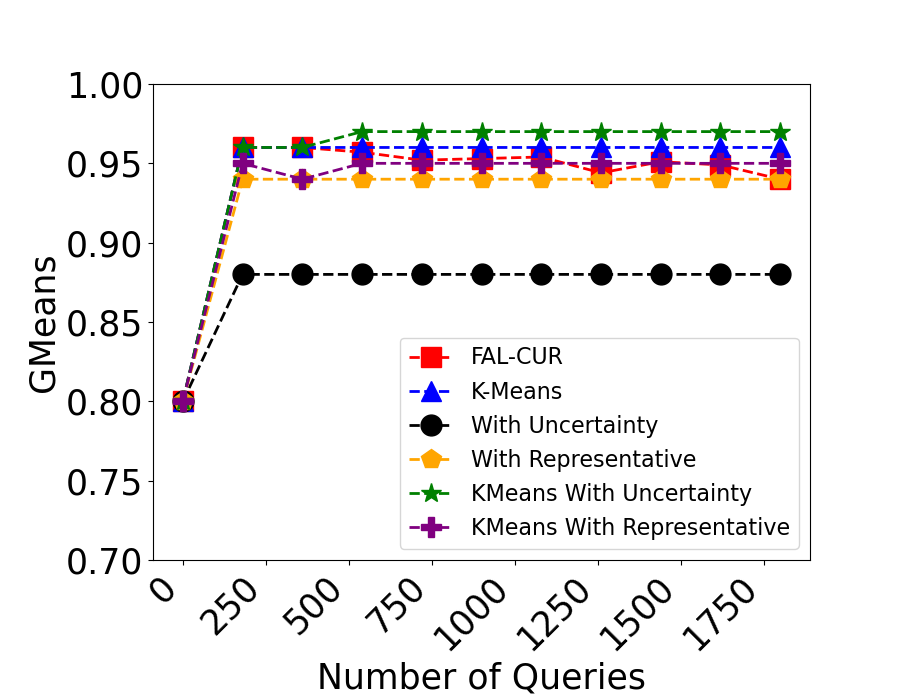}}  \hspace{0.13\textwidth}
\subfloat[Fairness]{\includegraphics[width = 0.4\linewidth]{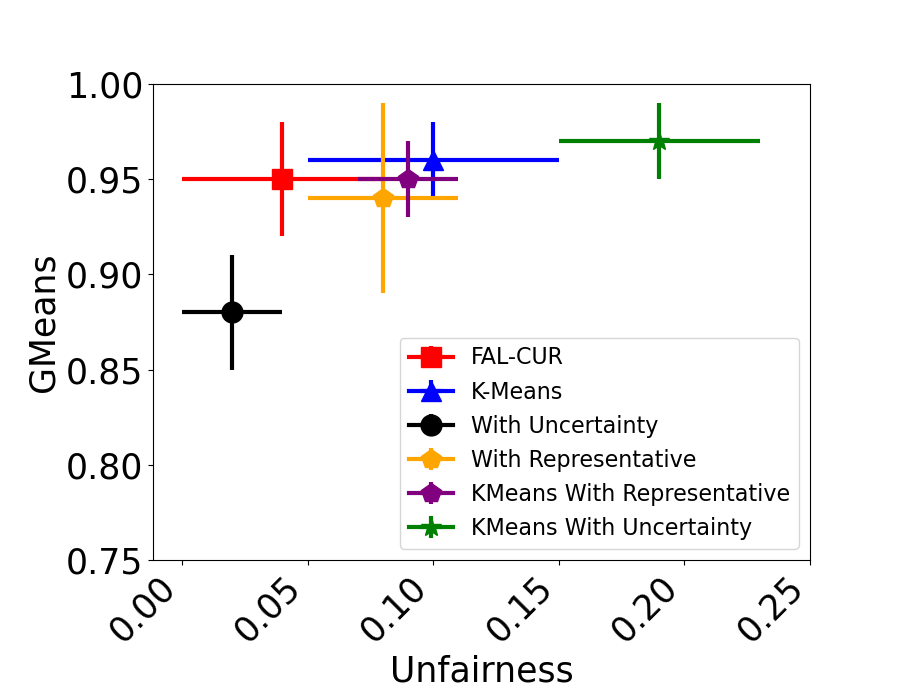}}

\caption{Ablation study on different variants using only fair clustering, uncertainty, and representativeness on Compass Dataset}\label{fig:ablation}
\end{figure*}

We perform an extensive ablation study with various versions of the FAL-CUR method to examine the effectiveness of its main components. These versions include: (i) using original KMeans instead of Fair Clustering, (ii) using only representative sampling in FAL-CUR by removing uncertainty sampling from the sample selection process, (iii) using only uncertainty sampling in FAL-CUR by removing representative sampling for sample acquisition, (iv) using KMeans with uncertainty, and (v) using KMeans with representative sampling. 
Figure \ref{fig:ablation} displays the accuracy and fairness (Statistical Parity) performance of each method. In terms of accuracy, standard KMeans clustering outperforms the Fair Clustering approach, particularly when combined with uncertainty sampling. However, using representative sampling causes a 0.05-point drop in the accuracy of KMeans clustering. On the other hand, from a fairness perspective, KMeans clustering with representative sampling yields better results than KMeans clustering with uncertainty sampling. As a result, to ensure good fairness performance without significantly compromising accuracy, FAL-CUR adopts the Fair Clustering approach instead of the standard KMeans clustering method.

The results indicate that eliminating fair clustering from the method enhances performance but decreases fairness. On the other hand, selecting uncertain samples with fair clustering boosts fairness scores but reduces performance, whereas using representative samples yields a consistently stable KMeans performance with a slightly reduced fairness score. Therefore, these findings emphasize the significance of all components - uncertainty and representative selection on fair clustering - which offer a reasonable trade-off between performance and fairness.

\begin{table}[t]
\centering
\caption {$\beta$ parameter experiments}
\begin{tabular}{l|l|l|l|l}
\hline \hline
$\beta$ parameters & SP   & E\_opp & E\_odd & GMeans \\ \hline \hline
1                  & 0.08 & 0.09   & 0.48   & 0.97  \\ \hline
0.9                & 0.07 & 0.08   & 0.45   & 0.96  \\ \hline
0.8                & 0.07 & 0.08   & 0.40   & 0.96  \\ \hline
0.7                & 0.06  & 0.07   & 0.39   & 0.96  \\ \hline
0.6                & 0.04 & 0.04   & 0.36   & 0.96  \\ \hline
0.5                & 0.03 & 0.03   & 0.36   & 0.93  \\ \hline
0.4                & 0.02 & 0.02   & 0.35   & 0.92  \\ \hline
0.3                & 0.02 & 0.02   & 0.35   & 0.92  \\ \hline
0.2                & 0.02 & 0.02   & 0.35   & 0.91  \\ \hline
0.1                & 0.02 & 0.02   & 0.35   & 0.9   \\ \hline 
0                  & 0.02 & 0.01   & 0.34   & 0.89  \\ \hline \hline
\end{tabular}\label{tab:beta}
\end{table}

\subsubsection{Analyzing the Impact of $\beta$}
Finally, we investigate the effect of $\beta$ parameter on the performance and fairness results. The $\beta$ value controls the balance between representative and uncertainty scores. A higher value of the $\beta$ parameter indicates that the model prefers representative over uncertainty. 
Table \ref{tab:beta} illustrates the influence of $\beta$ parameter on the performance result on the Compass dataset. From the table, one can infer that the performance, i.e., GMeans is increased when $\beta$ is higher; however, the unfairness metrics also increase. For example, when $\beta = 1$, the GMeans achieves the highest performance of 0.97 though the unfairness score is also the highest indicating the model is less fair to the sensitive class. However, when $\beta=0$, the model shows an improved fairness score with a decreased score in the performance. Our comprehensive experimental analysis recommends a $\beta$ value of 0.6; this not only achieves an impressive fairness score but also ensures the consistent performance of the model. It is also interesting to note that there is not much difference in fairness when $\beta$ is set to less than 0.4, though the GMeans shows a decreasing pattern. The ablation analysis results on other datasets are similar to the Compass dataset. 

\subsubsection{FAL-CUR under varying number of \textit{k}}
In this section, we examine the impact of fair-clustering on the FAL-CUR method. The underlying algorithm for the FAL-CUR is built on top of fair-clustering which is a KMeans clustering combined with fairness measures. Thus, it is interesting to investigate how the FAL-CUR method behaves under varying numbers of clusters denoted as \textit{k}. In this experiment, we use several values for \textit{k}, ranging from 50 to 250 while fixing the batch size as 180. 
Table \ref{tab:experiment_k} illustrates the results of this experiment on Compass dataset.
\begin{table}[htb!]
\centering
\caption {The performance of FAL-CUR for varying \textit{k} on Compass dataset.}
\begin{tabular}{l|l|l|l|l}
\hline \hline
k & SP   & E\_opp & E\_odd & GMeans \\ \hline \hline
50                  & 0.03 & 0.03   & 0.35   & 0.95  \\ \hline
100                & 0.03 & 0.03   & 0.36   & 0.96  \\ \hline
150                & 0.04 & 0.04   & 0.36   & 0.96  \\ \hline
180                & 0.04 & 0.04   & 0.36   & 0.96  \\ \hline
200                & 0.04  & 0.04   & 0.37   & 0.96  \\ \hline
250                & 0.05 & 0.05   & 0.37   & 0.96  \\ \hline
\hline \hline
\end{tabular}\label{tab:experiment_k} 
\end{table}

From the results, it can be observed that FAL-CUR consistently maintains a high level of performance across all \textit{k} values, with GMeans score consistently at 0.96 except for a very small deviation when $k=50$. Additionally, the statistical parity (SP), E\_opp and E\_odds values are also stable and do not fluctuate much as \textit{k} varies. The reason is that the fair-clustering method identifies fair clusters for a wide range of \textit{k} and it helps to maintain fairness in the active learning approach. These findings highlight the robustness of the FAL-CUR method for fair clusters with different sizes, showcasing its adaptability to various data distributions.

\subsection{Discussion}\label{sec:discussion}
The results show the effectiveness of the proposed FAL-CUR method in preserving fairness and maintaining accuracy for active learning tasks. To better understand the significance of these findings, we discuss them in further detail in this section as compared to other baseline methods in the active learning domain. 

First, as we discussed the proposed method performs better in terms of fairness. We, therefore,highlight the improvement of the proposed FAL-CUR method in terms of performance as compared to the best two state-of-the-art methods, i.e., FAL \cite{Anahideh2020FairAL} and ALOD-RE \cite{ALOD}. Table \ref{tab:percentage_improvement} shows the percentage of improvement of FAL-CUR as compared to FAL and ALOD-RE in terms of equalized odds. 
Overall, on average, FAL-CUR outperformed FAL by 18.20\% and ALOD-RE by 24.11\% based on the considered datasets.

\begin{table}[htbp]
\centering
\caption{Percentage Improvement of FAL-CUR with respect to two best baselines}\label{tab:percentage_improvement}
\begin{tabular}{c|cc}
\hline \hline
\multirow{2}{*}{\textbf{Dataset}} & \multicolumn{2}{c}{\textbf{\% Improvement}}                      \\ \cline{2-3} 
                         & \multicolumn{1}{c|}{FAL-CUR VS FAL} & FAL-CUR VS ALOD-RE \\ \hline
Adult                    & \multicolumn{1}{c|}{15.15}          & 9.68               \\ \hline
Compass                  & \multicolumn{1}{c|}{20.00}             & 5.26               \\ \hline
Loan                     & \multicolumn{1}{c|}{20.00}             & 33.33              \\ \hline
OULAD                    & \multicolumn{1}{c|}{17.65}          & 48.15              \\ \hline
\textbf{Average Improvement}      & \multicolumn{1}{c|}{18.20}          & 24.11              \\ \hline \hline
\end{tabular}
\end{table}
Secondly, EADA and ALOD-RE methods are recent additions to the active learning field, and both these methods select samples for labeling based on the likelihood of improving active learning accuracy. While EADA uses an energy-based approach to measure the uncertainty of each instance to be labeled, ALOD-RE combines uncertainty sampling with outlier detection for its sample selection mechanism. Both approaches were seen to improve AL accuracy, however often at the expense of fairness. 
In contrast, FAL, as the first method explicitly designed for fair active learning, demonstrated the ability to maintain fairness, albeit at the expense of accuracy. The FAL's performance is driven by the approach that centers around selecting samples based on their expected fairness score once labeled.
Furthermore, the proposed FAL-CUR method addresses this research gap by balancing both fairness and accuracy performance. This is achieved by employing fair clustering and a high-informative sample selection process, thus it is able to preserve fairness while keeping accuracy stable. This dual optimization of fairness and accuracy distinguishes FAL-CUR from other methods in the fair active learning task. 

The proposed method has several advantages over the most recent state-of-the-art methods, such as FAL-CUR's design, which combines fairness, uncertainty, and clustering, allowing modification of its core components and enhancing adaptability in various applications. 
Furthermore, in active learning tasks where fairness and accuracy are equally prioritized, FAL-CUR's sample acquisition process specifically selects samples that optimize both fairness and accuracy, leading to more efficient sample selection approach. 
Finally, While FAL-CUR exhibits promising potential, it has some limitations. We hypothesize that the performance of the proposed method is sensitive to the underlying data distribution and might falter when faced with different complex data distributions that one can look at further. 

\section{Conclusion and Future Work}\label{sec:conclusion}

\textcolor{black}{Active learning has proven to be an effective strategy for reducing annotation costs in machine learning. However, many existing active learning methods do not consider fairness concerns that may arise due to biased sample selection. This paper focuses on addressing the challenge of preserving fairness while maintaining stable accuracy in fair active learning tasks. 
The main contribution of this paper is a novel approach, called FAL-CUR, which ensures fairness in active learning. Specifically, FAL-CUR combines uncertainty and representative sampling with fair clustering, which has proven to be a more effective strategy for achieving fairness in active learning scenarios. Extensive experiments demonstrated that FAL-CUR surpasses the fairness performance of recent state-of-the-art fair active learning models. This result is supported by both the fair clustering and the acquisition function, which select samples leading to stable accuracy performance and fairness towards sensitive attribute samples. The ablation study revealed that both components – fair clustering and the acquisition function – are crucial to achieving this performance. Specifically, the study showed that fair clustering contributes to achieving fairness, while the acquisition function aids in selecting informative candidates for improved accuracy.} 
Finally, we would like to highlight several future directions that we intend to investigate. First, the proposed FAL-CUR approach is specifically designed for i.i.d. (Independent and Identically Distributed) data and extending it to non-i.i.d. data, such as graphs, can be a challenging task. Graph-structured data inherently possesses non-i.i.d. characteristics, where the contextual relationships between connected nodes contradict the assumption of independence in i.i.d. data. These dependencies contain valuable information that can significantly impact the learning process and, consequently, the model's performance. Thus, adapting the FAL-CUR algorithm to non-i.i.d. data requires adjusting its independence assumptions and introducing new mechanisms to account for inherent dependencies. Moreover, in its current form, FAL-CUR has been developed and tested in a batch learning context, where the entire dataset is available at once, and the model is not required to adapt to new data on the fly. Thus, it is interesting to investigate how the FAL-CUR method works in a data stream. Data stream learning deals with scenarios where data arrives continuously over time, requiring the model to update its knowledge incrementally and in real time. This presents unique challenges to fairness, as the distribution of the data can change over time and potentially leads to drifting or shifting biases.

\bibliographystyle{IEEEtranS}
\bibliography{references.bib}
\end{document}